\documentclass[10pt,twocolumn,letterpaper]{article}

\usepackage[pagenumbers]{iccv} 

\usepackage{graphicx}
\usepackage{float}
\usepackage{svg}
\usepackage{afterpage}
\usepackage{lipsum}

\definecolor{cvprblue}{rgb}{0.21,0.49,0.74}

\usepackage{array}
\usepackage{times}
\usepackage{epsfig}
\usepackage{graphicx}
\usepackage{float}
\usepackage{wrapfig}
\usepackage{amsmath,amssymb,amsthm}
\usepackage{algorithm,algorithmicx,algpseudocode}
\usepackage{bm,xspace}
\usepackage{comment}
\usepackage{multirow}
\usepackage{balance}
\usepackage{url}
\usepackage{booktabs}
\usepackage{etoolbox,siunitx}
\usepackage{calc}
\usepackage{pifont,hologo}
\usepackage{color}
\usepackage{adjustbox}
\usepackage{enumitem}
\usepackage[normalem]{ulem}  
\usepackage[pagebackref,breaklinks,colorlinks,linkcolor=red,citecolor=blue]{hyperref}
\usepackage{natbib}
\PassOptionsToPackage{square,numbers,comma,sort}{natbib}

\newcommand{\authorskip}{\hspace{3mm}}
\newcommand{\institutionskip}{\hspace{6mm}}

\definecolor{blue}{HTML}{0055cc}
\definecolor{red}{HTML}{cc1100}
\definecolor{orange}{HTML}{cc7700}
\definecolor{gray}{HTML}{efefef}
\definecolor{darkgreen}{rgb}{0.13, 0.55, 0.13}
\definecolor{darkgray}{HTML}{757575}

\renewcommand{\eqref}[1]{Eq.~\ref{#1}}

\newcolumntype{x}[1]{>{\centering\arraybackslash}p{#1}}
\newcolumntype{y}[1]{>{\raggedright\arraybackslash}p{#1}}
\newcolumntype{z}[1]{>{\raggedleft\arraybackslash}p{#1}}

\DeclareMathSymbol{@}{\mathord}{letters}{"3B}

\makeatletter
\DeclareRobustCommand\onedot{\futurelet\@let@token\@onedot}
\def\@onedot{\ifx\@let@token.\else.\null\fi\xspace}

\newcommand*{\Rom}[1]{\expandafter\@slowromancap\romannumeral #1@}
\newcommand*{\rom}[1]{\expandafter\romannumeral #1}

\def\1{\bm{1}}

\DeclareMathAlphabet{\mathsfit}{\encodingdefault}{\sfdefault}{m}{sl}
\SetMathAlphabet{\mathsfit}{bold}{\encodingdefault}{\sfdefault}{bx}{n}

\def\paperID{8026} 
\def\confName{ICCV}
\def\confYear{2025}

\title{ControlNeXt: Powerful and Efficient Control for Image and Video Generation}

\author{Bohao Peng\textsuperscript{\mdseries1} \authorskip 
Jian Wang\textsuperscript{1} \authorskip 
Yuechen Zhang\textsuperscript{1}\authorskip
Wenbo Li\textsuperscript{1}\authorskip 
Ming-Chang Yang\textsuperscript{1}\authorskip
Jiaya Jia\textsuperscript{1,2} 
 \\ 
\textsuperscript{1}CUHK \institutionskip
\textsuperscript{2}SmartMore \institutionskip \\
{\tt\small \url{https://github.com/dvlab-research/ControlNeXt}}
}
\date{}

\begin{document}


\twocolumn[{
\maketitle
\begin{center}
    \vspace{-.3in}
    \captionsetup{type=figure}
    \includegraphics[width=1.0\linewidth]{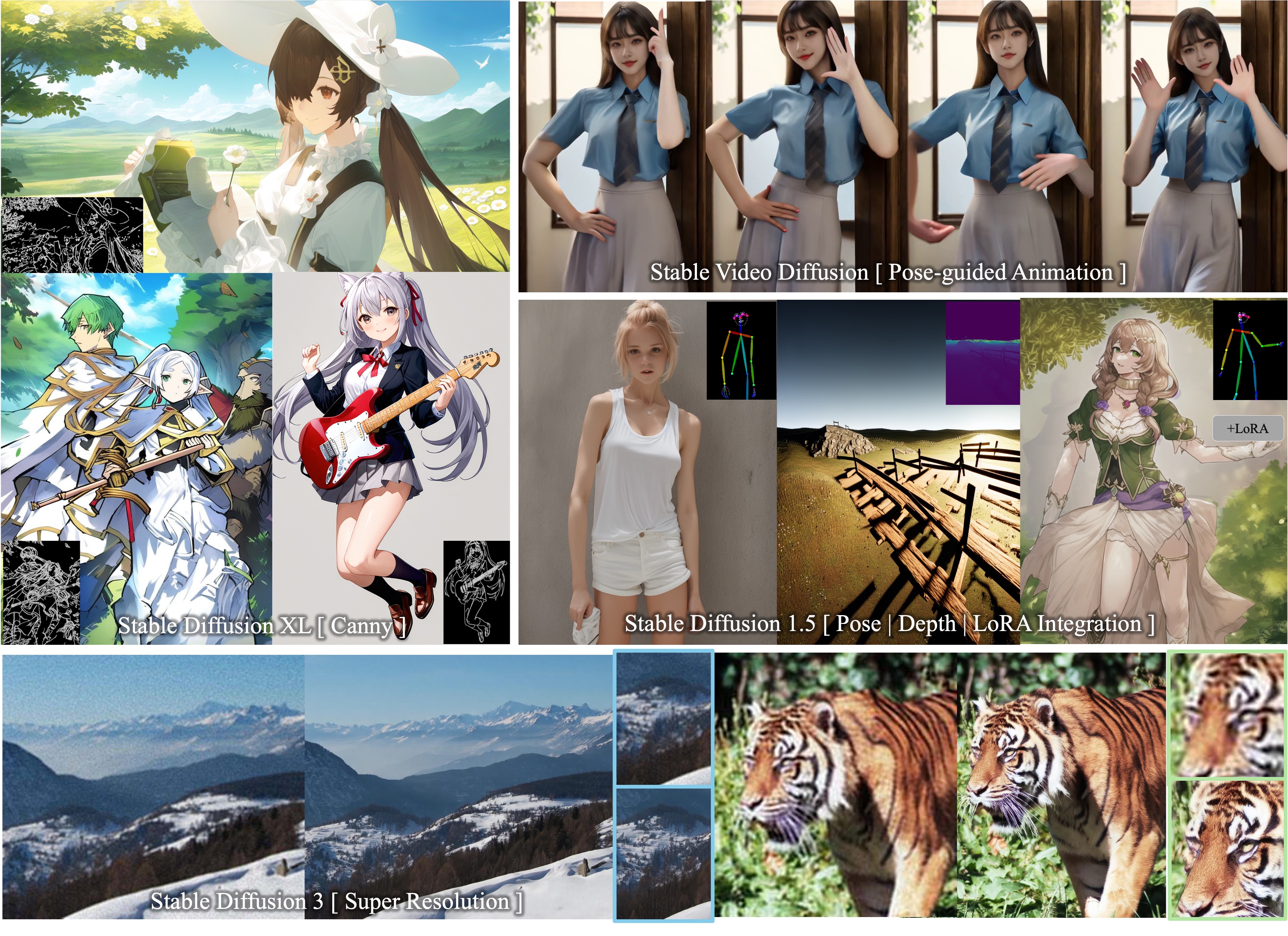}
    \vspace{-.3in}
    \captionof{figure}{ControlNeXt is a powerful and efficient method for controllable generation, emphasizing improved efficiency and generality. We demonstrate its applicability across diverse tasks and mainstream architectures. More results are provided in the supplementary materials.
    For more examples, please refer to our project page:~\url{https://pbihao.github.io/projects/controlnext/index.html}
    }
    \vspace{-.1in}
    \label{fig:summary}
\end{center}
}]


\begin{abstract}
    Diffusion models have achieved significant success in image and video generation, with conditional architectures such as ControlNet, Adapters, and ReferenceNet advancing spatial conditioning capabilities. However, existing controllable generation systems face limitations due to high computational requirements, slow convergence, and training instability, especially in resource-intensive video generation tasks. While researchers provide some task-specific solutions, they often lack flexibility and generality. To address these challenges, we introduce ControlNeXt—a powerful and efficient method for controllable image and video generation. ControlNeXt employs a lightweight architecture that integrates conditioning seamlessly, reducing learnable parameters by up to 90$\%$ compared to other approaches. Additionally, we propose Cross Normalization (CN), a stable and faster alternative to ``zero-convolution'' that improves training convergence. Extensive experiments across multiple models highlight ControlNeXt's generality and effectiveness in both image and video generation tasks.
\end{abstract}
\vspace{-.3in}
\section{Introduction}
\label{sec:intro}

Diffusion models generate complex, structured data by progressively refining a simple initial distribution, yielding realistic and high-quality results in image and video synthesis~\cite{rombach2022latentdiffusion, blattmann2023svd, guo2023animatediff, dhariwal2021diffusionbeatgans, lin2022collaborative, xing2025dynamicrafter}. Despite their success, these models often struggle with controllable generation, as achieving specific outcomes typically involves labor-intensive tuning of prompts and seeds. To address this, recent approaches~\cite{cao2024controllablesurvey, zhang2023controlnet, karras2023dreampose} incorporate auxiliary guidance signals, such as depth, pose skeletons, and edge maps, enabling more precise and targeted generation.

Popular controllable generation methods usually incorporate parallel branches or adapters to integrate control signals, as seen in ControlNet~\cite{zhang2023controlnet,wang2023disco}, T2I-Adapter~\cite{mou2024t2i}, and ReferenceNet~\cite{hu2024animateanyone}. These architectures process auxiliary controls in parallel while the main base model remains frozen. However, relying solely on the auxiliary components to capture controls always needs numerous parameters and introduces challenges, including increased computational demands, slower convergence, training instability, and limited controllability, as discussed in Secs~\ref{sec:training_convergence} and~\ref{sec:efficiency}. These issues are especially pronounced in resource-intensive video generation tasks. While T2I-Adapter~\cite{mou2024t2i} offers an efficient fine-tuning approach optimized for image generation, prioritizing efficiency often compromises controllability, rendering it less suitable for video generation and fidelity-oriented low-level tasks (details provided in the supplementary). Consequently, there is a pressing need for a controllable generation method that balances efficiency with general control capabilities.

This paper presents ControlNeXt, an efficient and general method for controllable generation, highlighting its enhanced performance across various tasks and backbone architectures (see Fig.~\ref{fig:summary}). Previous methods have demonstrated that control can be applied to pre-trained models~\cite{podell2023sdxl, blattmann2023svd, rombach2022latentdiffusion, zhang2024prompthighliter} by fine-tuning on small-scale datasets, suggesting that capturing control signals is not inherently difficult. Therefore, we argue that the base model itself is sufficiently powerful to be fine-tuned directly for controllability, without the need for additional auxiliary control components. This approach not only improves efficiency but also enhances the model's adaptability to complex tasks. To achieve this, we only use a lightweight convolution module to inject control signals, enabling the pre-trained model itself to learn controllable generation through selective fine-tuning. Specifically, we freeze most of the base model's parameters and selectively train a smaller subset, mitigating catastrophic forgetting~\cite{hu2021lora, peft, han2024parameter, chen2022vision} while significantly reducing training costs with minimal latency increase. This is especially crucial for complex tasks, such as video generation, where parameter-efficient fine-tuning (PEFT) methods like adapters and LoRA may fall short.

Furthermore, we introduce Cross Normalization as an alternative to Zero Convolution~\cite{zhang2023controlnet}, which serves as a ``bridge layer'' connecting the control branch to the base model. Zero Convolution, which initializes weights to zero, allows control signals to gradually influence the model during training. This approach is commonly used when fine-tuning pre-trained generation models, as introducing new components or parameters directly can lead to training collapse~\cite{zhang2023llamaadapter, xin2024parameter, karras2019style}. However, it also results in slow convergence as the learnable parameters initially struggle to receive the correct gradients. In this paper, we argue that training collapse primarily arises from the distributional mismatch between control guidance features and the intermediate features of the pre-trained model. This distributional dissimilarity makes the two sets of parameters incompatible. To address this, ControlNeXt introduces Cross Normalization, which aligns the data distributions, leading to more efficient and stable training and mitigating the ``sudden convergence'' problem observed in~\cite{zhang2023controlnet}.

We conduct a series of experiments on various generative backbones for image and video synthesis~\cite{sd1.5, rombach2022latentdiffusion, podell2023sdxl, blattmann2023svd}, demonstrating the generality and broad compatibility of ControlNeXt. Its lightweight design makes it a versatile, plug-and-play module that seamlessly integrates with other methods. Additionally, ControlNeXt accommodates LoRA weights~\cite{hu2021lora, ruiz2023dreambooth}, allowing for style modification without requiring further training. Our key contributions are summarized as follows: 
\begin{itemize}[leftmargin=5mm, itemsep=0mm, topsep=-1mm, partopsep=-1mm] 
\item We introduce ControlNeXt, a powerful and efficient method for controllable generation that strikes a balance between performance and general control capabilities. 
\item We propose Cross Normalization for fine-tuning large pre-trained models, enabling fast and stable convergence during training.
\item ControlNeXt serves as a lightweight, plug-and-play module that integrates seamlessly with LoRA weights to modify generation styles without additional training.
\end{itemize}

\section{Related Work}
\label{sec:related}

\noindent \textbf{Image and video diffusion models.} Diffusion probability models~\cite{ho2020denoising,song2019generative,dhariwal2021diffusionbeatgans} are advanced generative models that restore original data from pure Gaussian noise by learning the distribution of noisy data at various levels of noise. 
With their powerful capability to fit complex data distributions, diffusion models have excelled in several domains, including image and video generation. 
In the domain of image synthesis, diffusion models have demonstrably outperformed traditional Generative Adversarial Networks (GANs) in both image fidelity and diversity~\cite{dhariwal2021diffusionbeatgans}. As research in this field advances, diffusion models continue to push the boundaries of video generation, yielding unprecedented improvements in quality and temporal consistency. The predominant neural network architectures employed in diffusion models include UNet~\cite{ho2020denoising,dhariwal2021diffusionbeatgans,nichol2021glide,nichol2021improved} and DiT~\cite{peebles2023scalable}, with emerging alternatives such as U-ViT~\cite{bao2022all}.

In recent years, latent diffusion models have incorporated variational autoencoders (VAEs) to transfer the diffusion process to latent space, significantly accelerating the model's training and inference efficiency.
Leading image generation models like Stable Diffusion~\cite{sd1.5,rombach2022latentdiffusion,esser2024scaling,podell2023sdxl} have been widely adopted, used, and modified by the community.   
This success is attributed to streamlined and efficient model architecture designs, including sampling methods~\cite{meng2023distillation, song2023consistency,luo2023lcm}, the network structures of diffusion models~\cite{pernias2023wuerstchen}, and various additional and extended components. 

\noindent \textbf{Controllable generation.} Most recent models are guided by textual information as conditions~\cite{radford2021learning, devlin-etal-2019-bert, raffel2020exploring} to extract textual features that guide the generated content.
There are two main methods for introducing controllable conditions into image or video generation models: (i) training a large diffusion model from scratch to achieve controllability under multiple conditions~\cite{huang2023composer}, 
(ii) fine-tuning an adapter on a pretrained large model while keeping the original model parameters frozen~\cite{zhang2023controlnet, mou2024t2i}. Recent studies have attempted to control the outcomes of generative models by integrating additional neural networks into the foundation of diffusion models~\cite{zhang2024mimicmotion,xing2023make}.  
ControlNet guides image generation to align with control information by duplicating specific layers from pre-trained large models~\cite{zhang2023controlnet,wang2023disco}, but this approach introduces substantial parameters and latency. In contrast, the T2I-Adapter~\cite{mou2024t2i} employs an adapter for low-cost control, though it minimally affects original model, resulting in weaker control that limit its use for complex tasks.

\noindent 
\textbf{Distribution alignment in diffusion models.} Recent studies have highlighted the importance of distribution alignment~\cite{lu2021masa}. Zhang et al.\cite{zhang2024tackling} demonstrated that perturbing the initial noise distribution can mitigate generation issues by altering learned data distributions. In image-to-image translation and inpainting, techniques like\cite{lugmayr2022repaint} achieve improved results by aligning input noise with reference image distributions at intermediate stages. While Nie et al.\cite{chung2022diffusion} enhanced sample quality through posterior distribution alignment. These works collectively underscore the critical role of distribution matching in enhancing diffusion model performance and stability.

\begin{figure*}[ht!]
    \centering
    \includegraphics[width=1\linewidth]{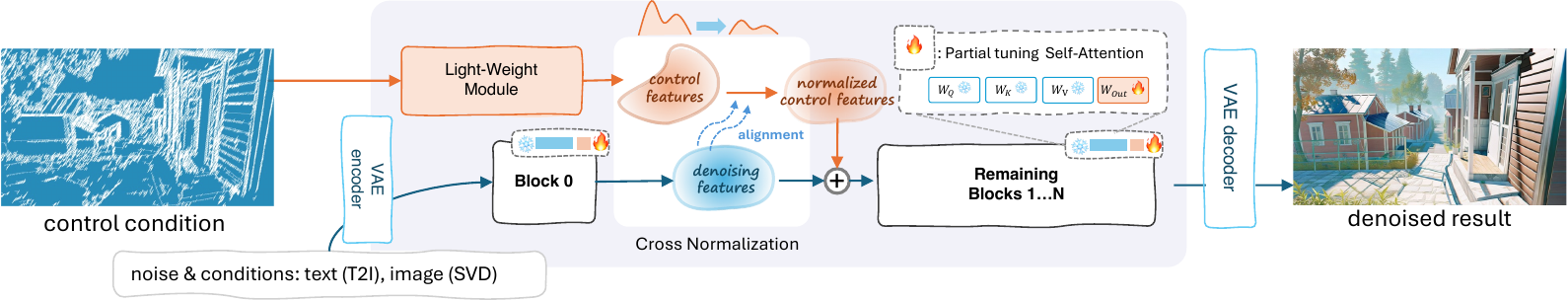}
    \caption{Training pipeline of ControlNeXt. We explore a powerful framework that achieves efficient controllable generation.}\label{fig:pipeline}
    \vspace{-.25in}
\end{figure*}

\section{Method}

In this section, we provide a detailed technical overview of ControlNeXt. We first introduce the necessary preliminaries for controllable generation in Sec.~\ref{sec:preliminary}. In Sec.~\ref{sec:architecture}, we delve into the analysis of the architecture design and prune it in order to make a concise and straightforward structure. Next, we introduce \textit{Cross Normalization} in Sec~\ref{sec:cross_norm}, which is designed for the efficient fine-tuning of large pre-trained models with additional components.

\subsection{Preliminaries}
\label{sec:preliminary}

Diffusion model (DM) is a type of generative model that generates data by reversing a gradual noise-adding process, transforming random noise into coherent data samples. The model's prediction for $x_t$ at time step $t$ depends only on $x_{t+1}$ and $t$:
\begin{equation}
    p_{\theta}(x_t|x_{t+1})=\mathcal{N}(x_t;\widetilde{\mu}_t,\widetilde{\beta}_tI),
\end{equation}
where $\theta$ represents the pre-trained model, $\widetilde{\mu}_t$ is the model's predicted target, and the variance $\widetilde{\beta}_t$ is computed from the posterior of forward diffusion:

\begin{equation}
    \widetilde{\beta}_t=\frac{1-\bar{\alpha}_{t-1}}{1-\bar{\alpha}_t}\beta_t.
\end{equation}

The loss function of diffusion models is the MSE loss function on the noise prediction $\hat{\epsilon}_\theta(x_t,t,c_t)$:

\begin{equation}
\mathcal{L}=w\cdot\mathbb{E}_{t,c_t,\epsilon\sim\mathcal{N}(0,1)}\Big[\Vert \epsilon-\hat{\epsilon}(x_t,t,c_t)\Vert^2\Big],
\end{equation}
where $c_t$ represents text prompts, and $w$ denotes the weight of the loss function. ControlNet~\cite{zhang2023controlnet} introduces controllable generation by integrating conditional control $c_f$. It calculates the loss function as:
\begin{equation}
    \mathcal{L}=w\cdot\mathbb{E}_{t,c_t,c_f,\epsilon\sim\mathcal{N}(0,1)}\Big[\Vert \epsilon-\hat{\epsilon}(x_t,t,c_t,c_f)\Vert^2\Big].
\end{equation}

\subsection{Architecture}
\label{sec:architecture}

\textbf{Motivation.} ControlNet~\cite{zhang2023controlnet} introduces a control branch to facilitate controllable generation, keeping the base model frozen to maintain its inherent generative quality. This branch, initialized as a replica of the original downsampling blocks, operates in parallel with the base model and employs a \textit{zero convolution} to integrate controls (more details provided in Sec.~\ref{sec:cross_norm}). Specifically:

\begin{equation}
    \bm{y}_c=\mathcal{F}_m(\bm{x})+\mathcal{Z}(\mathcal{F}_{cn}(\bm{x}, \bm{c};\Theta_{cn});\; \Theta_z),
\end{equation}
where $\mathcal{F}(\;\cdot\; ;\Theta)$ denotes a neural model with parameters $\Theta$, $\mathcal{Z}(\;\cdot\;; \Theta_z)$ indicates the \textit{zero convolution} layer. $\bm{x},\;\bm{y_c}\in\mathbb{R}^{h\times w\times c}$ and $\bm{c}$ are the 2D feature maps and conditional controls, respectively. 

Incorporating control capabilities in ControlNet entails significant computational costs. The additional branch increases considerable latency with extensive learnable parameters, particularly affecting video generation. The T2I-adapter~\cite{mou2024t2i} improves efficiency by replacing the control branch with an adapter. But this efficiency comes at the cost of reduced controllability and limits its effectiveness for complex tasks such as controllable video generation or low-level visual tasks. Moreover, freezing the base model and optimizing only the auxiliary modules limits overall model performance and slows convergence. To achieve general and efficient control, we propose allowing the pre-trained model to learn the control function directly.

\noindent
\textbf{Architecture.}  It is important to note that the pretrained model is typically trained on a large-scale dataset, such as LAION-5B~\cite{schuhmann2022laion}, whereas fine-tuning is always conducted on a much smaller dataset, often thousands of times smaller. Based on this, we assert that the pre-trained model is sufficiently powerful and general to capture controllability directly, without the need for heavy auxiliary components. 

We first eliminate the auxiliary components specifically designed to acquire control capabilities, such as ControlNet and adapters. To integrate the controls, we employ a compact convolution module only composed of multiple convolution blocks~\cite{he2016resnet}. Notably, this module is significant small and solely for extracting and aligning controls. For controllability, we propose directly fine-tuning a subset of the base model to enable it to capture guidance information. During training, we freeze most pretrained modules and optimize a small subset of parameters, such as the linear layers in the attention blocks. More details about the selected parts are provided in the supplementary material. Freezing most parameters also prevents catastrophic forgetting while maintaining training efficiency. And directly fine-tuning a subset of the base model is especially crucial for complex tasks like video generation, where PEFT methods such as LoRA and adapters~\cite{hu2021lora, wang2020k, liu2022few} may fall short. This approach enhances both effectiveness and efficiency, allowing adaptive adjustment of the learnable parameter scale to suit different tasks. Mathematically,
\begin{equation}
    \bm{y}_c = \mathcal{F}_m(\bm{x}, \mathcal{F}_c(\bm{c};\;\Theta_c)\;;\;\Theta_m'),
\end{equation}
where $\Theta_m' \subseteq \Theta_m$ represents a trainable subset of the pretrained parameters, and $\mathcal{F}_c$ is the lightweight convolution module. A more intuitive presentation is shown in Fig.~\ref{fig:pipeline}.

Regarding control injection, we aim to integrate control information at the earliest stage, allowing the base model to perceive the guiding information from the outset.
However, we found that directly adding the controls to the inputs results in training collapse, possibly due to the confusion and overlap between the controls and denoising features. Thus, we inject the controls after the first block, incorporating a residual connection to preserve the integrity of the main branch's identity transformation.  Further details will be provided in the supplementary. Controls are directly added to the denoising features after Cross Normalization introduced in Sec.~\ref{sec:cross_norm}, which further enhances training stability. Based on the above, ControlNeXt functions as a plug-and-play module, designed with a lightweight convolution and learnable parameters, represented as: \begin{equation}
    \mathcal{M}_c = \{\mathcal{F}_c(\;\cdot\;;\;\Theta_c),\;\Theta_d'\}
\end{equation}
where $\Theta_d'\subseteq\Theta_d$, and $\Theta_c <<\Theta_d$.

\subsection{Cross Normalization}
\label{sec:cross_norm}
\textbf{Motivation.} A key challenge in continual training of pretrained large models is how to appropriately introduce additional parameters and modules. Since directly combining new modules often leads to training collapse, recent works widely adopt zero initialization~\cite{zhang2023controlnet,zhang2023llamaadapter}, initializing the bridge layer that connects the based model and the added module to zeros. It ensures that newly introduced modules have no impact at the start of training, facilitating a stable warm-up phase. However, zero initialization also slows convergence and increases training challenges by preventing the modules from receiving accurate gradients at the start. This results in a phenomenon known as ``sudden convergence'' in controllable generation, where the model doesn't gradually learn the conditions but abruptly starts to follow them after an extended training~\cite{zhang2023controlnet}. 

\begin{figure*}[t]
    \centering
    \includegraphics[width=1.\linewidth]{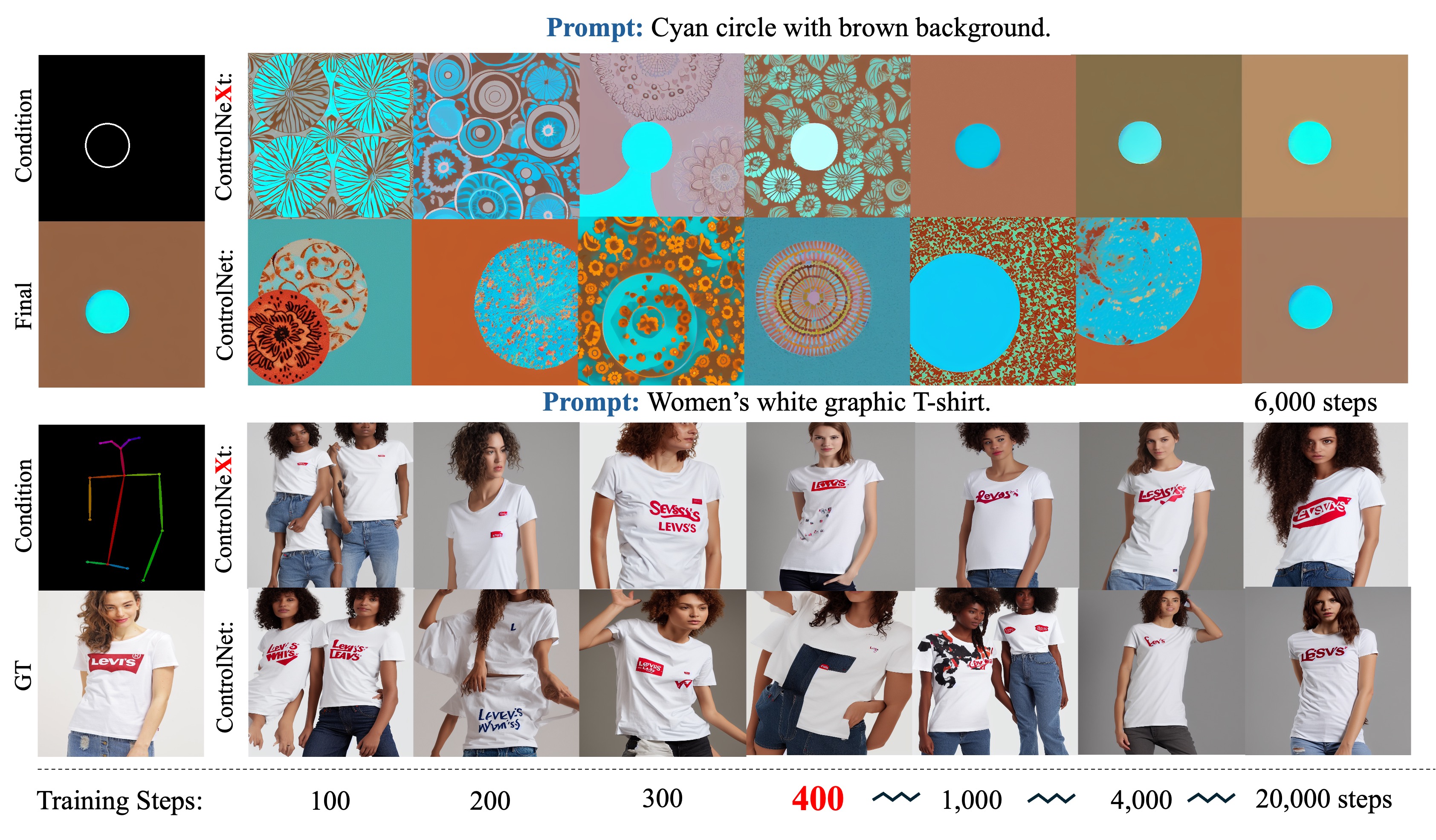}
    \vspace{-.3in}
    \caption{ControlNeXt achieves significantly faster training convergence and data fitting. It can learn to fit the conditional controls with fewer training steps, which significantly alleviates the \textit{sudden convergence} problem. }
    \vspace{-.3in}
    \label{fig:training_convergence} 
\end{figure*} 
\noindent
\textbf{Cross normalization.} The unaligned and incompatible data distribution among various features leads to training collapse and slow convergence~\cite{zhang2024tackling, lu2021masa,ioffe2015batch}. After training on the large-scale data, the pretrained generation model typically exhibits stable feature and distributions, characterized by consistent mean and standard deviation. However, the newly introduced neural modules are typically only initialized using random methods~\cite{he2019rethinking,kumar2017weight,lecun2015deep}, such as Gaussian initialization. 
This leads to the newly introduced neural modules producing feature outputs with significantly different data distributions, causing model instability when these outputs are directly added or combined.

Normalization techniques~\cite{ioffe2015batch,ba2016layer,wu2018group} standardize layer inputs, improving training stability and convergence. Inspired by these methods, we propose \textit{cross normalization} to align processed conditional controls with the main branch features, ensuring stable and efficient training.

We represent the feature maps processed by the base model and the lightweight convolution blocks as $\bm{x}_m$ and $\bm{x}_c$, respectively, where $\bm{x}_m, \bm{x}_c \in \mathbb{R}^{h \times w \times c}$. The key to Cross Normalization is to use the mean and variance calculated from the main branch $\bm{x}_m$ to normalize the control features $\bm{x}_c$, ensuring their alignment. First, calculate the channel-wise mean and variance of the denoising features,
\begin{equation}
    \bm{\mu}_{m} = \frac{1}{n} \sum_{i=1}^n \bm{x}_{m,i}\;,
\end{equation}
\begin{equation}
    \bm{\sigma}^2_{m} = \frac{1}{n} \sum_{i=1}^n (\bm{x}_{m,i} - \bm{\mu}_{m})^2\;,
\end{equation}
where $n=h\times w\times c$. Then, we normalize the control features using the mean and variance of the denoising features,
\begin{equation}
    \hat{\bm{x}}_c = \frac{\bm{x}_c - \bm{\mu}_m}{\sqrt{\bm{\sigma}^2_m+\varepsilon}} * \gamma,
\end{equation}
where $\varepsilon$ is a small constant added for numerical stability and $\gamma$ is a parameter that allows the model to scale the normalized value.  $\bm{x}_c=\mathcal{F}_c(\bm{c};\;\Theta_c)$ is the output control feature.

Cross Normalization aligns the distributions of the denoising and control features, serving as a bridge to connect the base model and control blocks. Our experiments in Sec.\ref{sec:training_convergence} show that this approach accelerates the training process, enabling the base model to capture guiding information from the outset. It facilitates early convergence and significantly alleviates the ``sudden convergence'' phenomenon.

\begin{figure*}[th!]
    \centering
    \includegraphics[width=1.\linewidth]{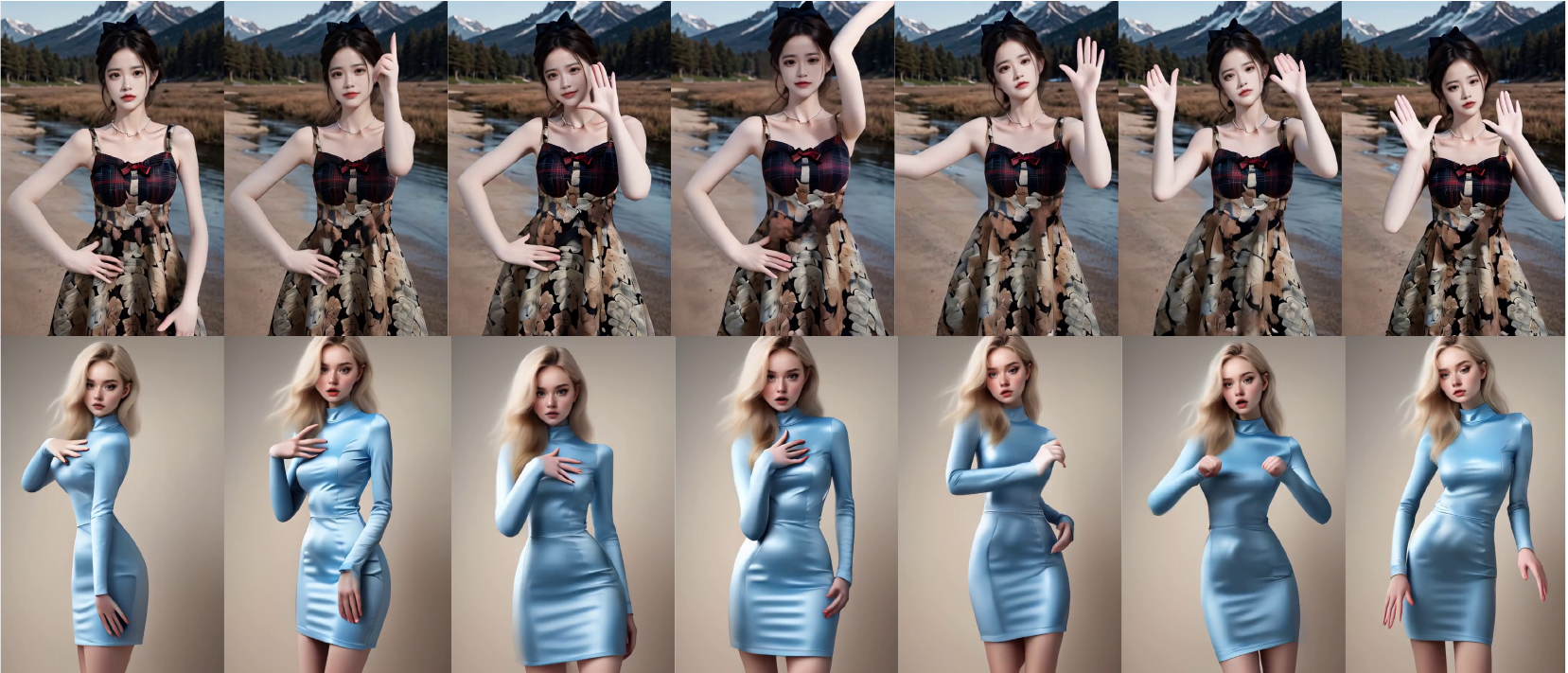}
    \vspace{-.2in}
    \caption{Detailed generation results of the stable video diffusion. We utilize the pose sequence as guidance for character animation.}
    \label{fig:svd_cases} 
    \vspace{-.2in}
\end{figure*}

\begin{figure*}[t]
    \centering
    \includegraphics[width=1.\linewidth]{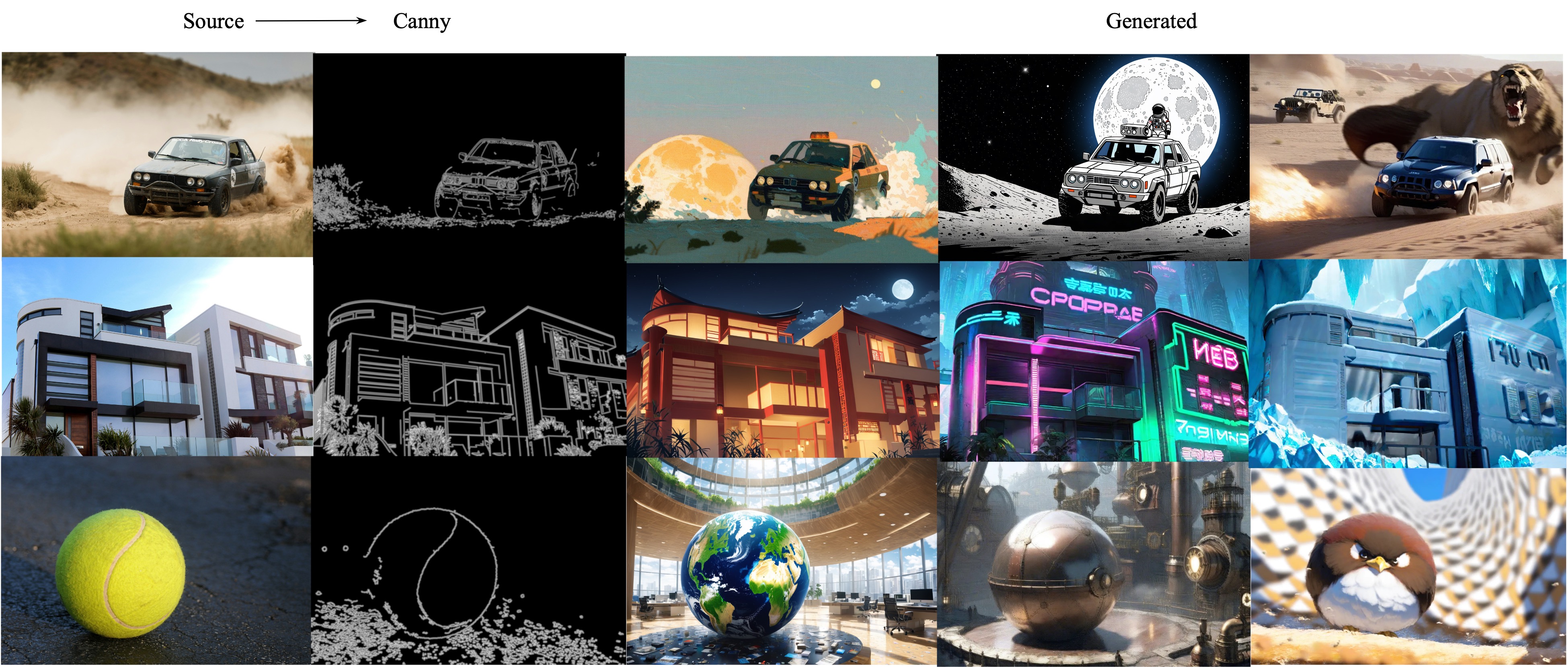}
    \caption{Detailed generation results of the stable diffusion XL are provided. We extract the Canny edges from the input image and implement the style transfer utilizing the SDXL model integrated with our proposed ControlNeXt framework.}
    \label{fig:sdxl_case} 
    \vspace{-.25in}
\end{figure*}

\section{Experiments}
\label{sec:experiments}

In this section, we present a series of experiments across various tasks and backbones. Our method exhibits exceptional efficiency and generality.

\begin{figure*}[t]
    \centering
    \includegraphics[width=1.0\linewidth]{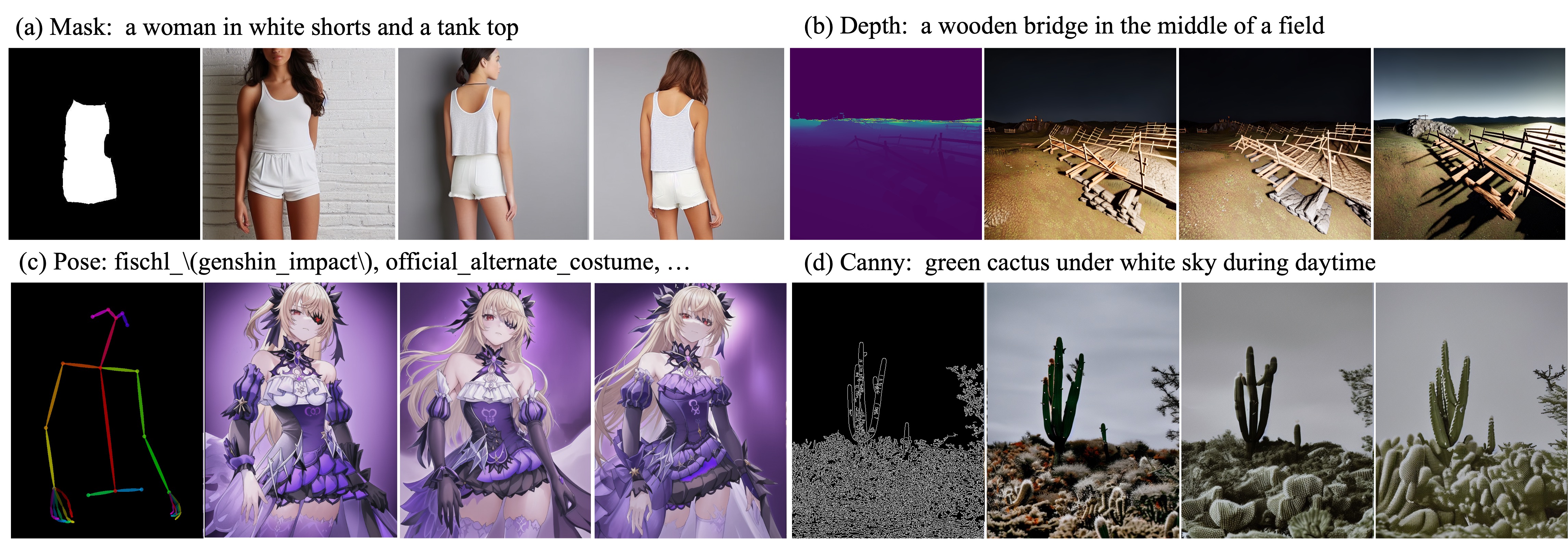}
    \vspace{-.3in}
    \caption{ControlNeXt supports various conditional controls types. We select ``mask'', ``depth'', ``pose'' and ``canny'', as the conditions.} 
    \label{fig:various_control_types} 
    \vspace{-.2in}
\end{figure*}

\subsection{Training Convergence}
\label{sec:training_convergence}

A typical problem for the controllable geneartion is the hard training convergence, which means that it requires thousands or more than ten thousands steps training to learn the conditional controls. This phenomenon, known as the \textit{sudden convergence} problem~\cite{zhang2023controlnet}, occurs when the model initially fails to learn the control ability and then suddenly acquires this skill. This is caused from such two aspects: 
\begin{enumerate}
    \item \textit{Zero convolution} inhibits the influence of the loss function, resulting in a prolonged warm-up phase where the model struggles to start learning effectively.
    \item The pretrained generation model is completely frozen, and ControlNet or the adapter cannot immediately affect the performance of the model.
\end{enumerate}
\noindent
In ControlNeXt, we eliminate these two limitations, resulting in significantly faster training. We conducted experiments using two types of controls, and the results and comparisons are shown in Fig.~\ref{fig:training_convergence}. It can be seen that ControlNeXt starts to converge after only a few hundred training steps, while ControlNet requires thousands of steps. ControlNeXt significantly alleviates the \textit{sudden convergence} problem.

\subsection{Generality}

To demonstrate the generality of our methods, we first apply our approach to various diffusion-based backbones, including Stable Diffusion 1.5~\cite{ho2020denoising,sd1.5}, Stable Diffusion XL~\cite{podell2023sdxl}, Stable Diffusion 3~\cite{esser2024scaling}, and Stable Video Diffusion~\cite{blattmann2023svd}. Our method covers a wide range of tasks, such as image generation, high-resolution generation, and video generation, utilizing various types of conditional controls. Qualitative results are shown in Fig.\ref{sec:intro}. Additionally, more generation results for stable video generation, where we use pose sequences as guidance for character animation, are presented in Fig.\ref{fig:svd_cases}. The results for SDXL are displayed in Fig.~\ref{fig:sdxl_case}, where we implement style transfer by extracting Canny edges from the input images and generating the output with our SDXL model. The results show that our method is adapting to various architectures and tasks.

\begin{table}[t]
    \centering
    \footnotesize  
    \setlength{\tabcolsep}{4.3pt}  
    \renewcommand{\arraystretch}{0.9}  
    \begin{tabular}{l|cc|cccc}
        \toprule
        \textbf{Metrics} & \multicolumn{2}{c|}{{Clip-score ($\uparrow$)}} & \multicolumn{3}{c}{{FID} ($\downarrow$)} \\
        \multirow{2}{*}{\textbf{Method}} & \multicolumn{2}{c|}{{Seg. Mask}} & {HED} & {MLSD} & {Pose} \\
        & ADE20K & COCO & COCO & COCO & COCO\\
        \midrule
        Gligen~\cite{li2023gligen} & 31.1  & - & 28.6  & - & 24.6  \\
        ControlNet~\cite{zhang2023controlnet} & 31.5  & 13.3  & 26.6  & 31.4  & 27.8  \\
        T2I-Adapter~\cite{mou2024t2i} & 30.6  & - & - & - & 29.6 \\
        ControlNet++~\cite{li2025controlnet++} & 31.9  & 13.3  & - & - & - \\
        Uni-Control~\cite{qin2023unicontrol} & 30.9  & - & \textbf{17.9}  & 26.2  & 26.6  \\
        \midrule
        $\textbf{ControlNeXt}_\text{(SD1.5)}$ & \textbf{32.7} & \textbf{29.5} & 20.4 & \textbf{21.1} & \textbf{23.0} \\
        \bottomrule
    \end{tabular}
    \vspace{-.1in}
    \caption{Comparison of different methods across metrics (Clip-score, FID) using Stable Diffusion 1.5 as the backbone.}
    \label{tab:qualitative_results}
    \vspace{-.1in}
\end{table}

\begin{table}[t]
	\centering
        \footnotesize
        \setlength{\tabcolsep}{3.5pt}  
	\renewcommand\arraystretch{1}
	{
		\begin{tabular}{c|cccccc}
			\toprule
			\textbf{{SR Method}} & {PSNR}$\uparrow$ & {SSIM}$\uparrow$ & {CLIPIQA}$\uparrow$ & {DISTS}$\downarrow$ & {MUSIQ}$\uparrow$ \\
			\midrule
                BSRGAN~\cite{zhang2021bsrgan} & 26.50 & 0.69 & 0.24 & 0.36 & 25.22 \\
 			SinSR~\cite{wang2024sinsr} & 26.83 & 0.64& 0.62 & 0.29 & 56.57\\
			SUPIR~\cite{yu2024supir} & 25.22 & 0.61& 0.56 & 0.26 & 59.02\\
            \midrule
			\textbf{Ours} (\textbf{SD3}) & \textbf{27.31} & \textbf{0.71} & \textbf{0.64}  & \textbf{0.20}  & \textbf{62.95} \\
			\bottomrule
		\end{tabular}}
    \vspace{-.1in}
    \caption{Quantitative results on super-resolution, evaluated with DRealSR~\cite{wei2020drealsr}, highlight Stable Diffusion 3 as the backbone.}
    \vspace{-.1in}
    \label{tab:super_revolution}
\end{table} 

\begin{table}[h]
    \centering
    \footnotesize
    \setlength{\tabcolsep}{3pt} 
    \begin{tabular}{l|ccccccc}
    \toprule
    \textbf{Method}         & MagicPose~\cite{chang2023magicpose}   & MuseV~\cite{musev}  & Mimic~\cite{zhang2024mimicmotion} & $\text{ControlNeXt}$  \\ 
    \midrule
    FVD $(\downarrow)$        & 916 & 754    & 594 & \textbf{576} \\ 
    \bottomrule
    \end{tabular}
    \vspace{-.1in}
    \caption{Comparison of character animation performance~\cite{jafarian2021tiktokdance} using stable video diffusion.}
    \vspace{-.3in}
    \label{tab:qualitative_results_svd}
\end{table}


\noindent
\textbf{Various conditional controls.} ControlNeXt also supports various types of conditional controls. In this subsection, we choose ``mask'', ``depth'', ``pose'' and ``canny'' as the conditional controls, shown in Fig.~\ref{fig:various_control_types} from top to bottom, respectively. All the experiments are constructed based on the Stable Diffusion 1.5 architecture~\cite{sd1.5}. \\
\noindent
\textbf{Quantitative Resuls.} Tab.\ref{tab:qualitative_results} show a quantitative comparison on ADE20K and COCO~\cite{lin2014microsoft, zhou2019semantic} with Stable Diffusion 1.5 as backbone. ControlNeXt achieves state-of-the-art results with efficiency and generality. For the super-resolution task, we conduct experiments using the DRealSR benchmark~\cite{wei2020drealsr} with Stable Diffusion 3 as the backbone in a DiT-based architecture. Results in Tab.~\ref{tab:super_revolution} highlight our advancements. Following prior work~\cite{zhang2024mimicmotion}, we evaluate video generation on the character animation task using the TikTok dataset~\cite{jafarian2021tiktokdance}. Results are presented in Tab.~\ref{tab:qualitative_results_svd}. Furthermore, we apply our method to the DiT-based video generation backbone~\cite{hong2022cogvideo,yang2024cogvideox, lin2024open_sora}, Open-Sora-Plan, for video outpainting tasks, following the setup of prior work~\cite{fan2023m3ddm_sdm}. The results are shown in Tab.~\ref{tab:video_outpainting}.

\begin{table}[t]
	\centering
        \footnotesize
        \setlength{\tabcolsep}{2.3pt}
	\renewcommand\arraystretch{1}
	{
		\begin{tabular}{@{}c|ccc|ccc@{}}
			\toprule
			\multirow{2}{*}{\textbf{{Method}}}&\multicolumn{3}{c|}{DAVIS dataset~\cite{ren2020davis}}&\multicolumn{3}{c}{YouTube-VOS~\cite{duarte2019youtube}}\\
		& {PSNR}$\uparrow$ & {SSIM}$\uparrow$  & {FVD}$\downarrow$  & {PSNR}$\uparrow$ & {SSIM}$\uparrow$ & {FVD}$\downarrow$ \\
			\midrule
 			SDM~\cite{fan2023m3ddm_sdm} & 20.02 & 0.7078  & 334.6 & 19.91 & 0.7277 & 94.8 \\
			M3DDM~\cite{fan2023m3ddm_sdm} & 20.26 & 0.7082 & 300.0 & 20.20 & 0.7312 & 66.6\\
            \midrule
			\textbf{Ours}(Open-Sora-Plan) & \textbf{20.33} & \textbf{0.7576} & \textbf{290.7}  & \textbf{20.23} & \textbf{0.7661}  & \textbf{60.3} \\
			\bottomrule
		\end{tabular}}
	\vspace{-.1in}
    \caption{Video outpainting task with Open-Sora-Plan as backbone.}
    \label{tab:video_outpainting}
    \vspace{-.1in}
\end{table} 

\begin{table}
    \centering
    \footnotesize
    \setlength{\tabcolsep}{3.5pt}
    \begin{tabular}{l|cc|cc|cc}
        \toprule
        \multirow{2}{*}{Backbone} & 
        \multicolumn{2}{c|}{ControlNet} & 
        \multicolumn{2}{c|}{ControlNeXt (Ours)} & 
        \multicolumn{1}{c}{Base Model} \\
        & Total & Learnable & Total & Learnable & Total   \\
        \midrule
        SD1.5  & 1,220 & 361 & 865 & \textbf{30}  & 859  \\
        SDXL    & 3,818 & 1,251 & 2,573 & \textbf{108}  & 2,567  \\
        SVD     & 2,206 & 682 & 1,530 & \textbf{55} & 1,524  \\
        \bottomrule
    \end{tabular}
    \vspace{-.1in}
    \caption{Comparison of the total and learnable parameters of different methods with various backbones.}
    \label{tab:nums_params}
    \vspace{-.15in}
\end{table}

\begin{figure}[t]
    \centering
    \includegraphics[width=1.0\linewidth]{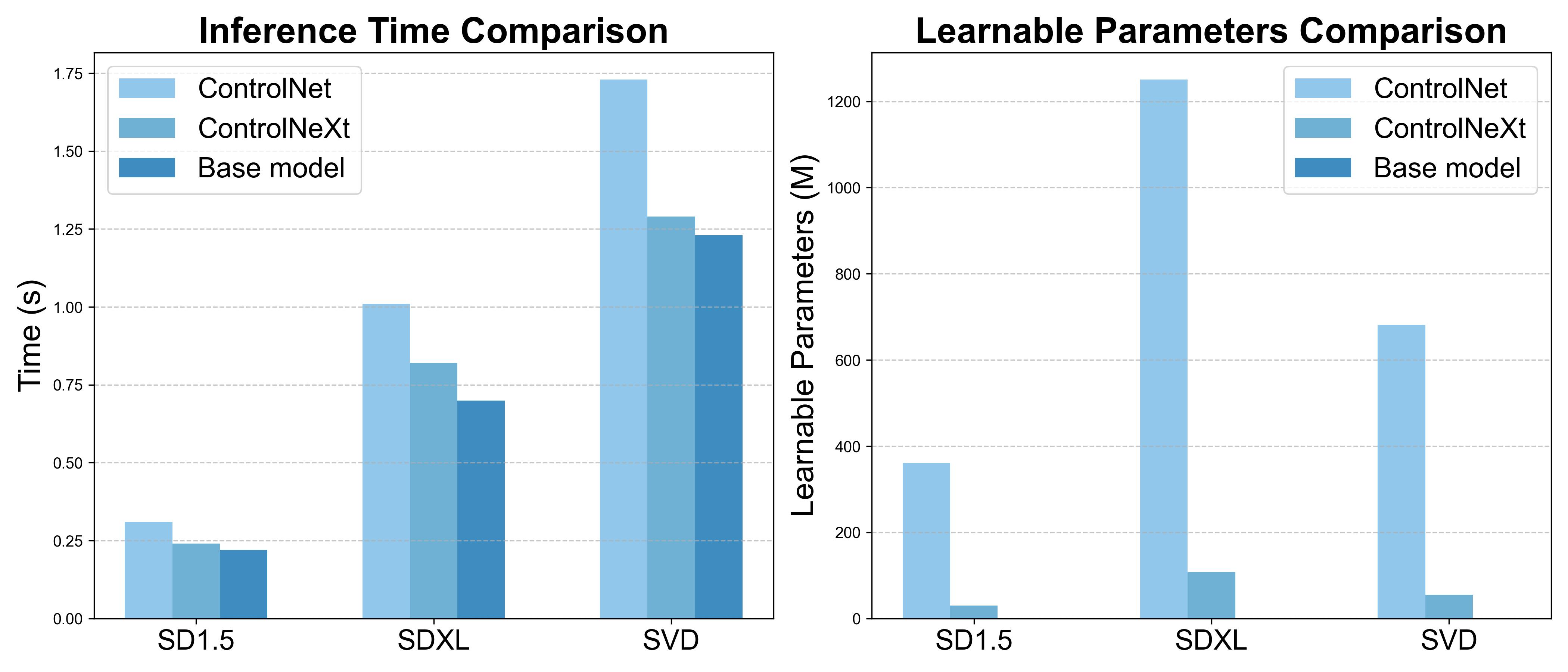}
    \vspace{-.3in}
    \caption{Efficiency comparisons of ControlNeXt.}
    \label{fig:Efficiency} 
    \vspace{-.2in}
\end{figure}

\subsection{Efficiency}
\label{sec:efficiency}
In this section, we compare the efficiency of various backbones, focusing primarily on comparison with ControlNet~\cite{zhang2023controlnet} for its representativeness and generalizability. Alternatives such as T2I-Adapter~\cite{mou2024t2i} are limited to image generation and lack support for all tasks and backbones. Further details are provided in the supplementary materials. A comprehensive comparison is shown in Fig.~\ref{fig:Efficiency}.\\

\begin{figure}[t]
    \vspace{-.2in}
    \centering
    \includegraphics[width=1.0\linewidth]{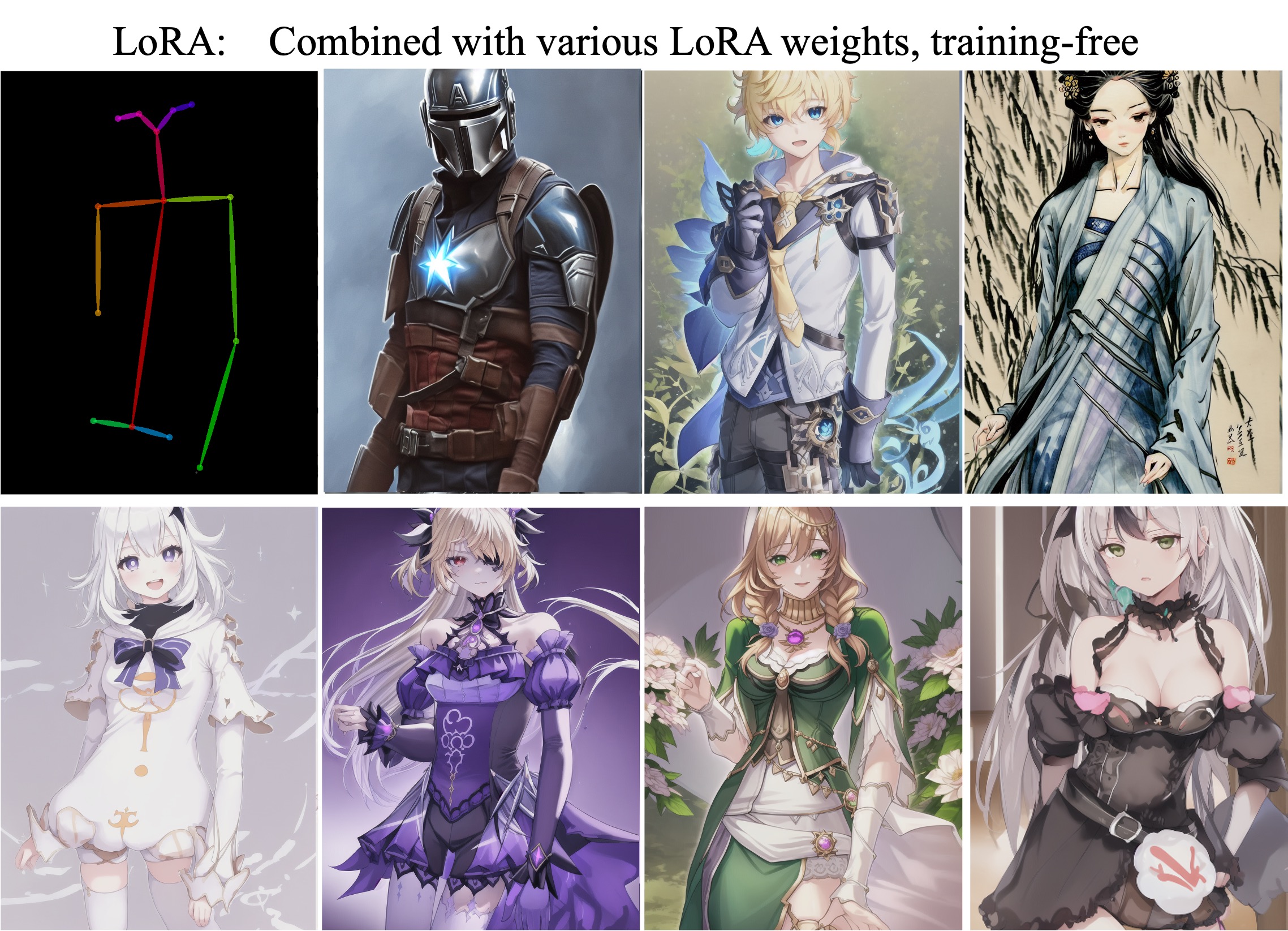}
    \vspace{-.2in}
    \caption{Our method can serve as a plug-and-play module that adapts to various LoRA weights with training-free.}
    \vspace{-.1in}
    \label{fig:plug_and_play}
\end{figure}

\begin{figure}[t]
    \vspace{-.1in}
    \centering
    \includegraphics[width=1.0\linewidth]{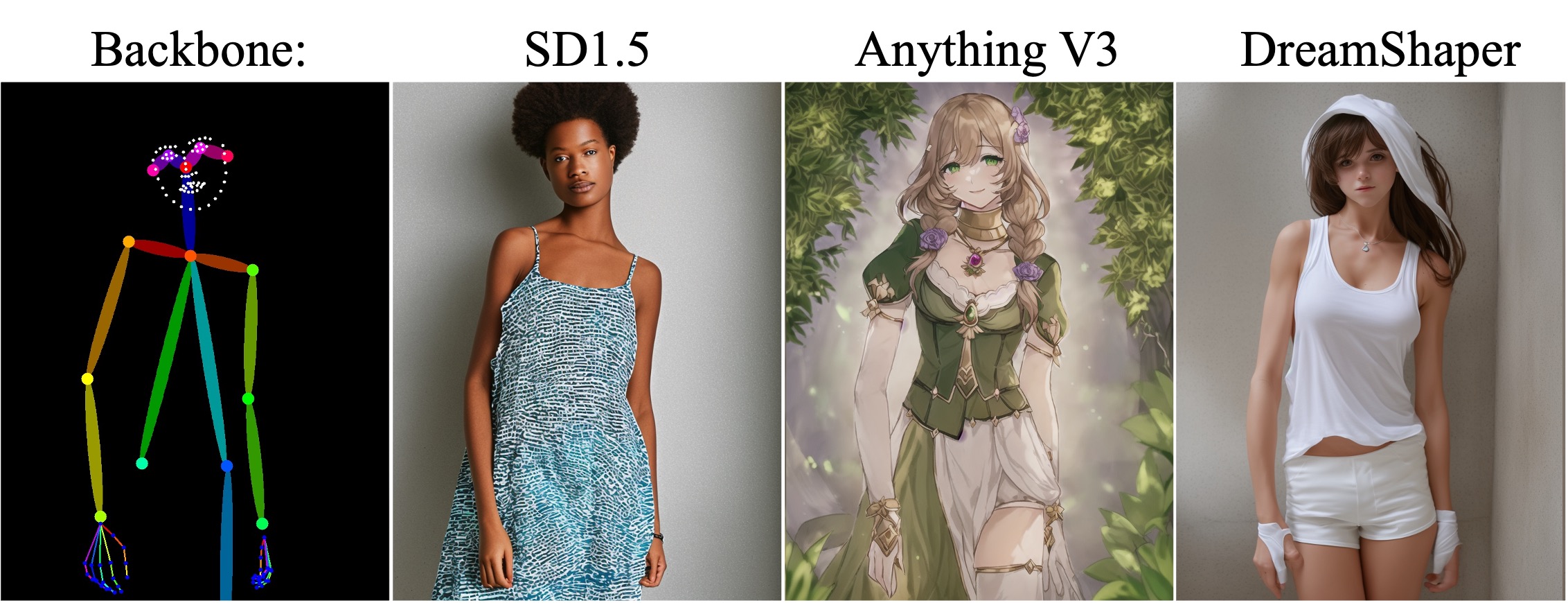}
    \vspace{-.3in}
    \caption{ControlNeXt is compatible with a variety of backbones.}
    \label{fig:various_backbone}
    \vspace{-.3in}
\end{figure}

\noindent
\textbf{Parameters.} We present statistics on the parameters, including the total and learnable parameters, calculated only for the UNet model (excluding the VAE and encoder parts). And the results are shown in Tab.~\ref{tab:nums_params}. It can be seen that our method only adds a lightweight module with minimal additional parameters, maintaining consistency with the original pretrained model. As for training, our method requires at most less than 10\% of the learnable parameters. You can also adaptively adjust the amount of learnable parameters for various tasks and performance requirements.\\

\noindent
\textbf{Inference time.} We compare the inference time of different methods with various base models. The results are shown in Tab.~\ref{tab:inference_latency}, which presents the computational time of one inference step, considering only the UNet and ControlNet. It can be seen that our method increases latency minimally compared to the pretrained base generation model. This ensures outstanding efficiency advantages for our method.

\subsection{Additional Studies}

\begin{figure}[t]
    \centering
    \vspace{-.2in}
    \includegraphics[width=1.0\linewidth]{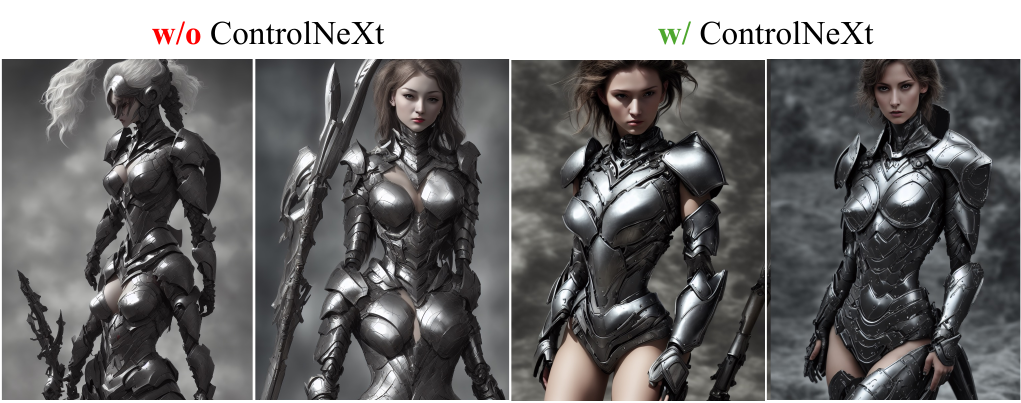}
    \vspace{-.3in}
    \caption{ControlNeXt serving as a plugin-unit to ensure a stable generation with minimal costs.}
    \label{fig:stable_generation} 
\end{figure}

\noindent
\textbf{Training free integration.} We first collected various LoRA weights downloaded from Civitai~\cite{civitai}, encompassing diverse generation styles. We then construct experiments on various backbones, including SD1.5~\cite{sd1.5}, AnythingV3~\cite{Anythingv3} and DreamShaper~\cite{DreamShaper}. The results are shown in Fig.~\ref{fig:plug_and_play} and Fig.~\ref{fig:various_backbone}. It can be observed that ControlNeXt can integrate with various backbones and LoRA weights in a training-free manner, effectively altering the quality and styles of generated images. It also facilitates stable generation with minimal effort and cost as shown in Fig.~\ref{fig:stable_generation}. We use a simple text prompt, \textit{i.e.}, ``one girl,'' with the `pose' condition, enabling high-quality generation without detailed textual descriptions.\\

\begin{figure}[t]
	\vspace{-.1in}
		\centering
		\includegraphics[width=1.\linewidth]{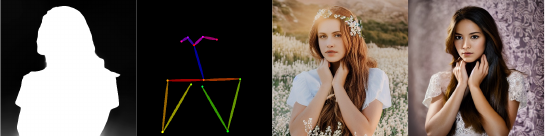}
	\vspace{-.2in}
    \caption{Controllable generation under multiple conditions.}
    \label{fig:multiple_conditions}
    \vspace{-.1in}
\end{figure}

\noindent
\textbf{Multiple conditions.} We fine-tune lightweight control modules for `depth' and `pose' conditions and integrate them into the main branch without other operations. Learnable parameters in the main branch are assigned to non-overlapping blocks for each condition. Results in Fig.~\ref{fig:multiple_conditions} demonstrate our method's support for multiple conditions.

\noindent
\textbf{Information intergration.} We conduct ablation studies to validate our method's effectiveness, with results in Tab.~\ref{tab:ablation_cn}. Beyond qualitative improvements, our approach allows adjustable control impact on the backbone, including turning off the control signal—unlike direct concatenation or cross-attention. This is especially beneficial when combined with classier-free-guidance for optimal results.

\begin{table}[t]
    \centering
    \footnotesize
    \setlength{\tabcolsep}{8pt}
    \begin{tabular}{l|lll|c}
        \toprule
        \multirow{2}{*}{Method}&\multicolumn{3}{c|}{Inference Time (s)} \\
        & SD1.5 & SDXL& SVD&$\Delta$\\ 
        \midrule
        ControlNet&0.31&1.01&1.73&\textsubscript{+} 41.9\% \\
        $\textbf{ControlNeXt}_{(\text{Ours})}$&\textbf{0.24}&\textbf{0.82}&\textbf{1.29}&\textsubscript{+} \textbf{10.4\%} \\
        \midrule
        Base model&0.22&0.70&1.23&- \\
        \bottomrule
    \end{tabular}
    \vspace{-.1in}
    \caption{Comparison of inference time with various backbones. }
    \label{tab:inference_latency}
    \vspace{-.1in}
\end{table}

\begin{table}[t]
	\centering
        \footnotesize
        \setlength{\tabcolsep}{4pt}
	\renewcommand\arraystretch{1}
	{
		\begin{tabular}{c|cccccc}
			\toprule
			 {FID}$\downarrow$/Clip$\uparrow$& {\textbf{CrossNorm }} & {{Concat}} & {{CrossAttn}} & ControlNet \\
			\midrule
 			HED & \textbf{20.4} / \textbf{29.4} & 27.2 / 28.9  & 271 / 22.7 & 26.6 / -\\
			MLSD & \textbf{21.1} / \textbf{29.2}& 24.5 / 28.8 & 381 / 20.6 & 31.4 / -\\
			\bottomrule
		\end{tabular}}  
        \vspace{-.1in}
        \caption{Comparative analysis of different methods for integrating conditional controls.}
        \label{tab:ablation_cn}
        \vspace{-.35in}
\end{table} 

\vspace{-.1in}
\section{Conclusion}
\label{sec:conclusion}

This paper presents ControlNeXt, an advanced and efficient method for controllable image and video generation. ControlNeXt employs a compact architecture, eliminating heavy auxiliary components to minimize latency overhead and reduce trainable parameters. We propose \textit{Cross Normalization} for finetuning pre-trained large models, improving training convergence in both speed and stability. Extensive experiments across various image and video generation backbones demonstrate the effectiveness and generality.


\clearpage
{
    \small
    \bibliographystyle{ieeenat_fullname}
    \bibliography{main}
}

\clearpage
\begin{figure*}[h]
    \centering
    \includegraphics[width=.9\linewidth]{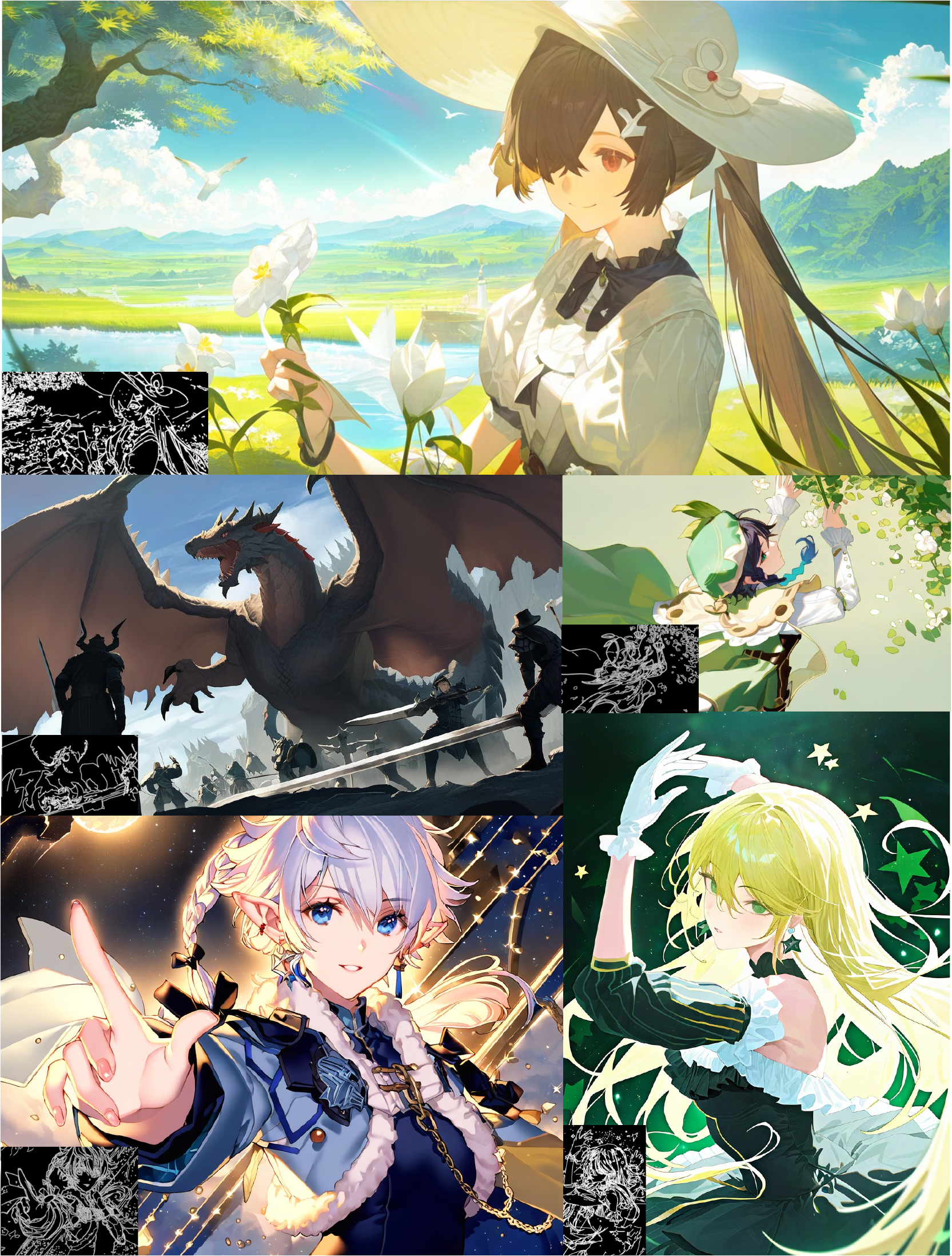}
    \caption{Stable Diffusion XL.}
    \label{fig:sdxl_cases} 
    \vspace{-.2in}
\end{figure*}
\clearpage
\begin{figure*}[h]
    \centering
    \includegraphics[width=.9\linewidth]{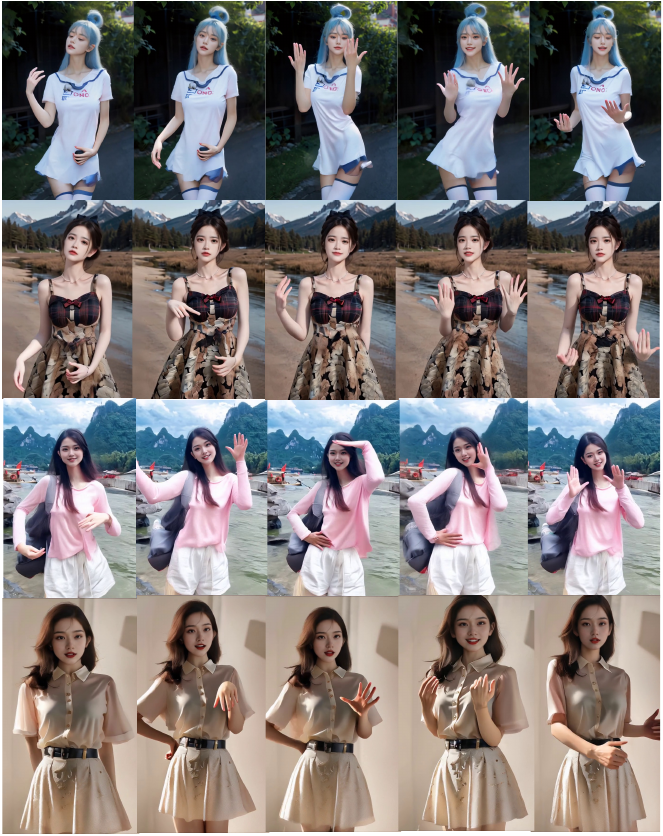}
    \caption{Stable video diffusion.}
    \label{fig:svd} 
    \vspace{-.2in}
\end{figure*}
\clearpage
\begin{figure*}[h]
    \centering
    \includegraphics[width=.8\linewidth]{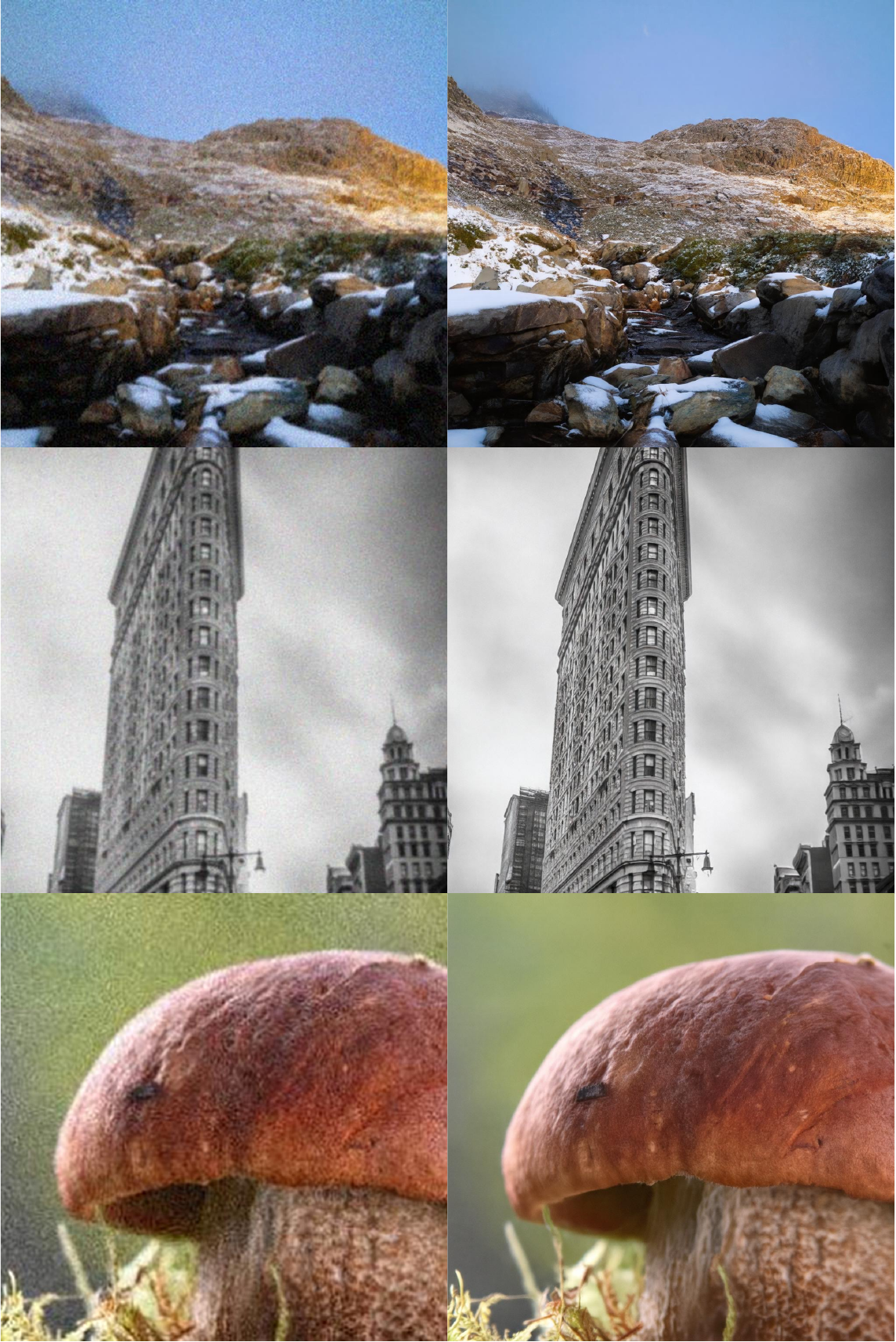}
    \caption{Stable Diffusion 3.}
    \label{fig:sd3_cases} 
\end{figure*}

\begin{figure*}[t]
  \centering
  \includegraphics[width=.9\linewidth]{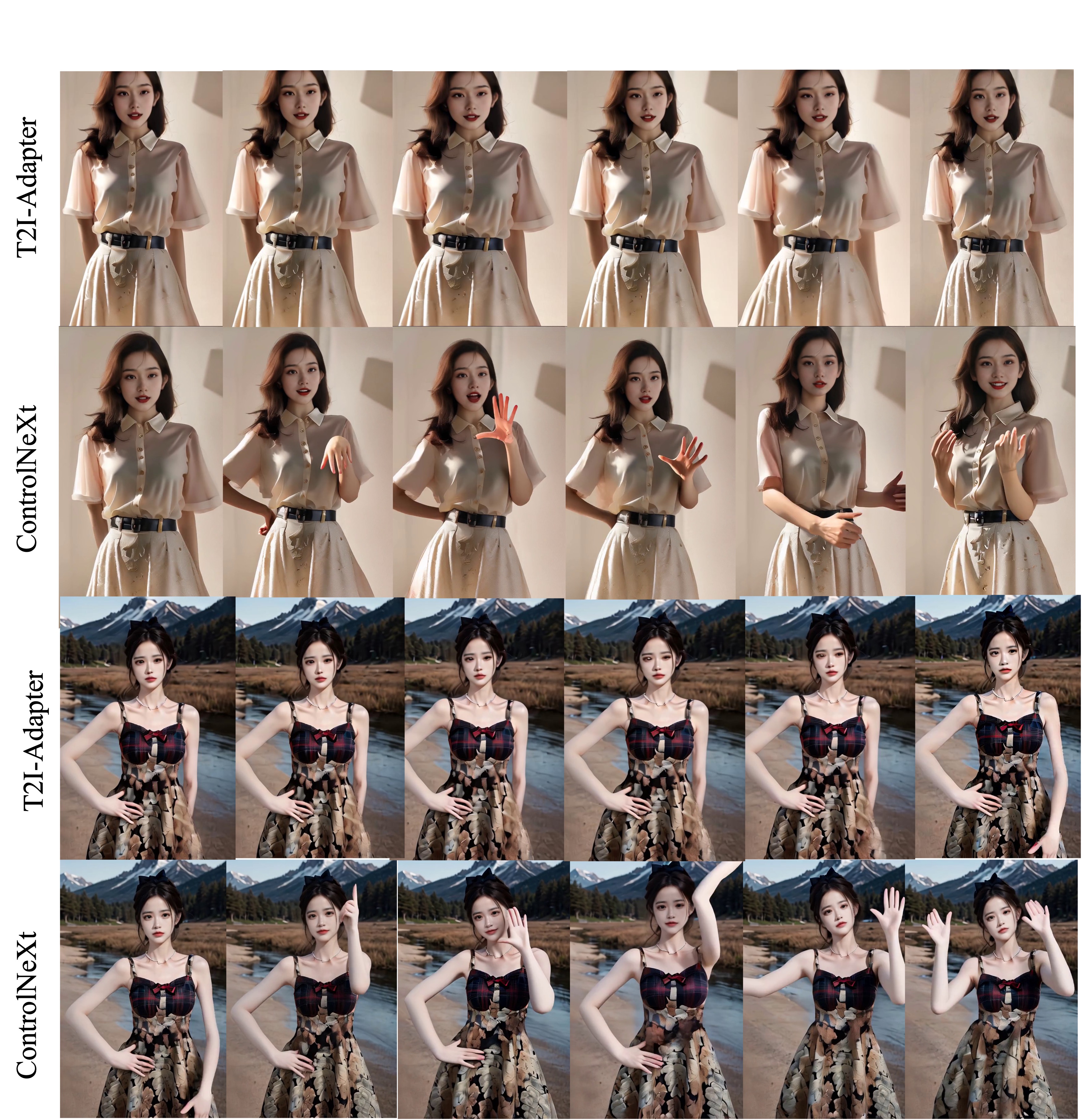}
  \caption{Comparison with the T2I-adapter. The T2I-adapter is specifically designed for image generation and is challenging to adapt for more complex tasks, such as video generation.}
  \label{fig:vis_compare_t2i}
\end{figure*}


\end{document}